\documentclass{article}

\usepackage{arxiv}

\usepackage[utf8]{inputenc} 
\usepackage[T1]{fontenc}    
\usepackage[numbers]{natbib}
\usepackage{url}            
\usepackage{booktabs}       
\usepackage{amsfonts}       
\usepackage{nicefrac}       
\usepackage{microtype}      
\usepackage{lipsum}		
\usepackage{graphicx}
\usepackage{natbib}
\usepackage{doi}

\title{Feature Selection Library (MATLAB Toolbox)}

\date{January, 2024}	

\author{ \href{https://orcid.org/0000-0003-4170-914X}{\includegraphics[scale=0.06]{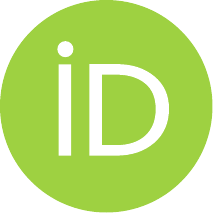}\hspace{1mm}Giorgio Roffo}\\
	Department of Computer Science\\
	Italy \\
	\texttt{giorgio.roffo@neural-brain.io}  \\
}

\renewcommand{\headeright}{FSLIB}
\renewcommand{\undertitle}{Technical Report}
\renewcommand{\shorttitle}{FSLIB}
\usepackage{array}
\newcolumntype{C}[1]{>{\centering\arraybackslash}p{#1}}
\hypersetup{
pdftitle={A template for the arxiv style},
pdfsubject={q-bio.NC, q-bio.QM},
pdfauthor={David S.~Hippocampus, Elias D.~Striatum},
pdfkeywords={First keyword, Second keyword, More},
}

\begin{document}
\maketitle

\begin{abstract}
The Feature Selection Library (FSLib) introduces a comprehensive suite of feature selection (FS) algorithms for MATLAB, aimed at improving machine learning and data mining tasks. FSLib encompasses filter, embedded, and wrapper methods to cater to diverse FS requirements. Filter methods focus on the inherent characteristics of features, embedded methods incorporate FS within model training, and wrapper methods assess features through model performance metrics. By enabling effective feature selection, FSLib addresses the curse of dimensionality, reduces computational load, and enhances model generalizability. The elimination of redundant features through FSLib streamlines the training process, improving efficiency and scalability. This facilitates faster model development and boosts key performance indicators such as accuracy, precision, and recall by focusing on vital features. Moreover, FSLib contributes to data interpretability by revealing important features, aiding in pattern recognition and understanding. Overall, FSLib provides a versatile framework that not only simplifies feature selection but also significantly benefits the machine learning and data mining ecosystem by offering a wide range of algorithms, reducing dimensionality, accelerating model training, improving model outcomes, and enhancing data insights.
\end{abstract}

\keywords{Feature Selection Library \and Feature Selection \and Computer Vision}

\section{Introduction}
Feature selection (FS) is a critical preprocessing step in machine learning, aimed at enhancing the performance and efficiency of classification models by reducing data dimensionality and eliminating noise. This process is instrumental in refining the dataset to include only the most informative features, thereby improving the accuracy and computational efficiency of classifiers. The essence of FS lies in its ability to filter out redundant or irrelevant information, a task that becomes increasingly challenging as data complexity grows. Traditionally, feature identification has been manually conducted by experts within various learning domains. However, this manual selection process may not always identify the most relevant or non-redundant features, highlighting the necessity for automatic feature selection methods. These automated processes are adept at discarding irrelevant, redundant, and noisy data, which is crucial for enhancing learning and classification performance. The importance of feature selection is well-acknowledged across numerous fields, including machine learning, artificial intelligence, computer vision, and data mining, finding application in information retrieval, user re-identification, recommendation systems, and visual object tracking, among others \cite{KristanLMFPCVHL16,roffo2016object,RoffoBMVC2016,vinciarelli2018regular,roffo2016object}.

Within the FS domain, methodologies are broadly classified into three primary categories: filter methods, wrapper methods, and embedded methods. Filter methods prioritize the intrinsic properties of the data, operating independently of any classifier to select features. Wrapper methods involve optimizing a predictor as part of the selection process, often yielding superior results compared to filter methods but at a higher computational cost\cite{Roffo5555,Roffo2013,Cristani:2012,Roffo2014,Roffo:icmi2014,RoffoEMPIRE16,DBLP:conf/eccv/2016w2}. Embedded methods integrate the feature selection process with the model learning mechanism, making the two inseparable. This approach is unique in its interaction between feature selection and learning, exemplifying the importance of the function class under consideration in determining feature relevance. 

Wrapper methods employ classifiers to assess the utility of feature subsets, embedded methods integrate selection within the classifier learning process, and filter methods analyze data's intrinsic properties irrespective of the classifier \cite{guyon2006feature}. These approaches can execute two primary operations: ranking and subset selection. Ranking evaluates the importance of individual features without considering their interactions, whereas subset selection identifies the optimal feature combination for use. Although these operations can be sequential or exclusive, subset selection is inherently supervised, contrasting with the potentially unsupervised nature of ranking \cite{Guyon:2002,Bradley98featureselection,Grinblat:2010,Hutter:02feature,liu2008,LeiYi10.1109,Quanquanjournals}.

Embedded methods uniquely intertwine feature selection with the learning algorithm, making them inseparable from the learning process and thereby tailoring feature selection to the specific model in use, such as Support Vector Machines \cite{guyon2006feature}. This integration of selection and learning processes within embedded methods is particularly relevant in the context of modern attention and self-attention mechanisms in transformers. These mechanisms inherently perform a form of feature selection by weighting the importance of different parts of the data (e.g., tokens in natural language processing), thus aligning with the principles of embedded feature selection methods but on a data-driven, adaptive basis.

Feature selection is recognized as an NP-hard problem \cite{guyon2006feature}, necessitating the use of heuristic or suboptimal search strategies to manage the combinatorial complexity of selecting an optimal subset from a potentially vast feature space. Filters approach this challenge by initially ranking features individually before extracting a subset, with notable examples including Inf-FS \cite{Roffo_2015_ICCV}, EC-FS \cite{RoffoECML16}, MutInf \cite{Hutter:02feature}, and Relief-F \cite{liu2008}. Conversely, wrapper and embedded methods evaluate and select feature subsets in a more holistic manner, with FSV \cite{Bradley98featureselection,Grinblat:2010} and SVM-RFE \cite{Guyon:2002} serving as key examples. Given the NP-hard nature of feature selection, with the combinatorial explosion of feature subsets to evaluate, efficient and heuristic search strategies are employed \cite{guyon2006feature}. Filters prioritize individual features through ranking before extracting a subset, exemplified by methods such as Inf-FS and MutInf, while wrappers and embedded techniques evaluate and select subsets directly, with notable examples including FSV and SVM-RFE \cite{Roffo_2015_ICCV,RoffoECML16,Hutter:02feature,liu2008,Bradley98featureselection,Grinblat:2010,Guyon:2002,andersoncvpr}.

\begin{figure}[!]
\centering
\includegraphics[width=0.6\linewidth]{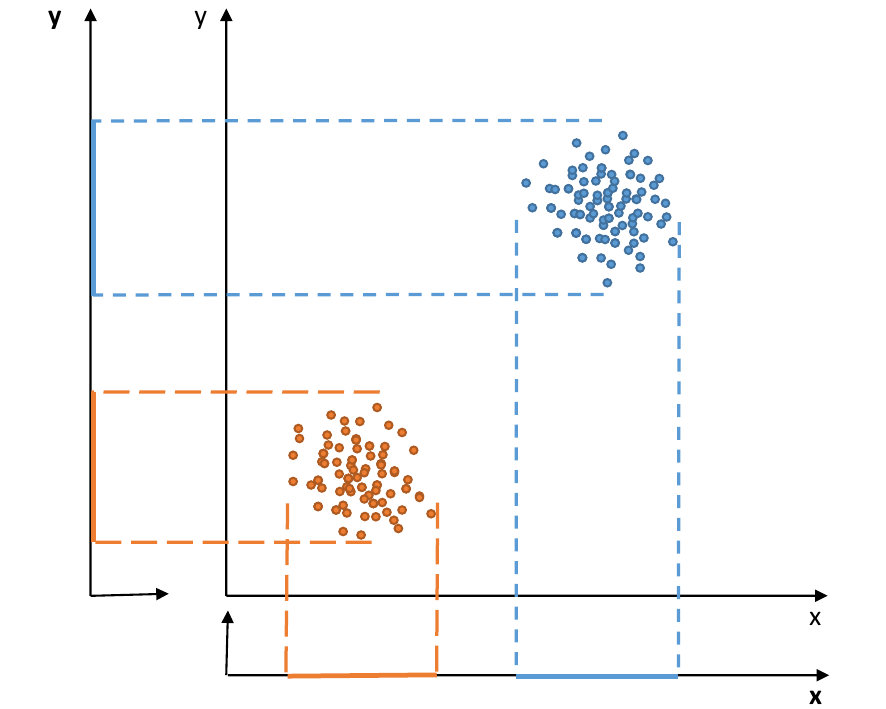}
\includegraphics[width=0.6\linewidth]{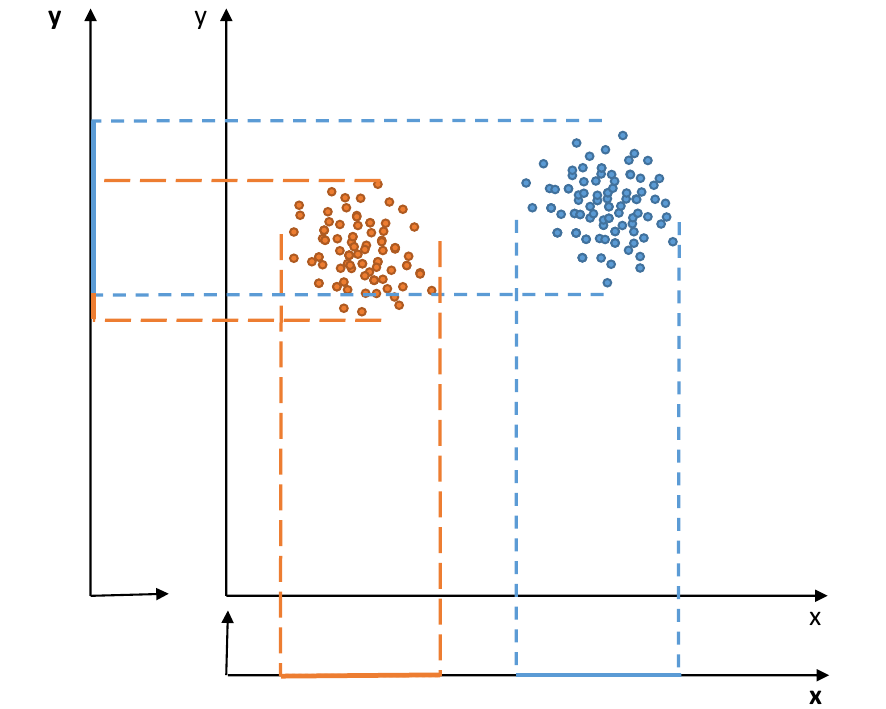}
\caption{(Left) In this example, features x and y are redundant, because feature x provides the same
information as feature y with regard to discriminating the two clusters. (Right) In this example, we consider feature y to be irrelevant, because if we omit x, we have only one class, which is uninteresting.}\label{fig:redundant}
\end{figure}
 
Feature selection algorithms are also distinguished by their learning context, being either supervised or unsupervised. Supervised algorithms leverage class labels to retain features with high predictive relevance to the classes while eliminating redundant or irrelevant features. This is based on the premise that a high-quality feature subset is one where features are both predictive of the class and non-redundant with each other \cite{Gennari:1989,roffofeature}. Unsupervised learning, devoid of class labels, presents a more challenging scenario for feature selection, emphasizing the need for innovative approaches to identify relevant features without explicit class information.

This report is structured to provide an extensive examination of existing feature selection methods, followed by a detailed exploration of the proposed toolbox's contributions. By situating our toolbox within the broader context of attention mechanisms and transformer models, we underscore its relevance and potential impact on current and future developments in machine learning and artificial intelligence.

The organization of this report is structured to provide a comprehensive review of existing feature selection methods, followed by an in-depth exploration of the contributions made by the proposed toolbox. Through this examination, we aim to underscore the toolbox's relevance and potential to advance the field of machine learning, particularly in the context of attention mechanisms and transformer models, thereby enriching the toolkit available for researchers and practitioners alike. A brief overview of the existing methods is given in Section~\ref{sec:SoA}. Finally, conclusions are provided in Section~\ref{sec:conc}.

\section{Feature Selection Techniques}\label{sec:SoA} 

The objective of feature selection (FS) is to identify a subset of features that optimizes a model's generalization ability, effectively reducing the risk associated with prediction. This section delves into various FS strategies, highlighting their methodologies, applications, and inherent challenges.

Relief-F \cite{liu2008} exemplifies a supervised, iterative approach to FS, assessing feature quality by their capacity to distinguish between neighboring samples. Its effectiveness notwithstanding, Relief-F's performance is contingent upon data availability and does not adequately address feature redundancy.

SVM-RFE \cite{Guyon:2002}, an embedded FS method, systematically eliminates features using a backward elimination strategy, prioritizing those that significantly contribute to sample separation via a linear SVM. Despite its strengths, SVM-RFE shares Relief-F's limitations in handling redundant features.

The Fisher method \cite{Quanquanjournals} represents a fast, filter-based FS strategy, calculating feature scores based on the ratio of interclass separation to intraclass variance. Similarly, mutual information (MI) \cite{Hutter:02feature} evaluates features based on the mutual information between the distribution of feature values and class membership. Both methods independently assess features before aggregating the top-ranked ones for final selection.

Infinite Latent Feature Selection (ILFS) \cite{roffo2017infinite,roffofeature2} introduces a generative, probabilistic approach to FS, bypassing the combinatorial challenge by considering all possible feature subsets. This method prioritizes features based on a graph weighted by PLSA-inspired mixing weights, highlighting the concept of feature relevancy.

Feature selection via Eigenvector Centrality (EC-FS) \cite{RoffoECML16,obertino2016infinite} employs an affinity graph where features are nodes, and node importance is determined through Eigenvector Centrality. This filter method underscores the importance of a feature as a function of its neighbors' importance, identifying central nodes as candidate features for effective classification.

Addressing the challenge of unsupervised learning scenarios, Inf-FS \cite{Roffo_2015_ICCV,RoffoBMVC2016} and the Laplacian Score (LS) \cite{HCN05a} utilize graph-based methods to assess feature importance. Inf-FS assigns importance scores by considering all possible feature subsets as paths on a graph, whereas LS evaluates features based on their ability to preserve local geometric structures through a nearest neighbor graph.

Recent unsupervised wrapper methods like the Dependence Guided Unsupervised Feature Selection (DGUFS) \cite{JunGuo_AAAI_2018_DGUFS} and Feature Selection with Adaptive Structure Learning (FSASL) \cite{du2015unsupervised} highlight the significance of data interdependencies in FS. DGUFS optimizes feature subset selection by balancing the dependence on original data and cluster labels, while FSASL integrates structure learning with FS, albeit with considerable computational demands for high-dimensional data.

The Least Absolute Shrinkage and Selection Operator (LASSO) \cite{liu2007computational} incorporates a regularization process to minimize prediction error and select features with significant coefficients, exemplifying a method where FS is a consequence of model optimization.

Lastly, the feature selection via concave minimization (FSV) \cite{Bradley98featureselection} innovatively integrates FS into the training of an SVM through linear programming, demonstrating a wrapper approach that directly ties the FS process to model training efficacy.

\subsection{The Feature Selection Library (FSLib) Approaches} 

The Feature Selection Library (FSLib), available on File Exchange through MATLAB Central by MathWorks, encompasses a comprehensive suite of algorithms designed for feature selection. These algorithms are categorized based on their operational approach and learning context, facilitating a structured methodology for selecting the most relevant features from datasets.

In Table~\ref{table:compmethods}, the algorithms within FSLib are classified according to two primary dimensions: their \textit{type} and \textit{class}. The \textit{type} of an algorithm indicates its methodological approach to feature selection, including:

\textit{f} for filters, which prioritize statistical measures of feature relevance independent of any machine learning model's performance,
\textit{w} for wrappers, which evaluate feature subsets based on the performance of a specific predictive model, and
\textit{e} for embedded methods, which integrate the feature selection process directly into the model training, making the selection of features and the training of the model interdependent.
The \textit{class} of an algorithm refers to its learning context, distinguishing between:

\textit{s} for supervised methods, which utilize labeled training data to guide the feature selection process, aiming to improve prediction accuracy by focusing on features most relevant to the output variable, and
\textit{u} for unsupervised methods, which do not rely on labeled data, instead identifying features based on the intrinsic structure of the data itself, such as clustering or density estimation.
Furthermore, the table documents the computational complexity of each algorithm, where available in the literature. Computational complexity is a critical factor in the selection of an appropriate feature selection method, as it determines the algorithm's scalability and feasibility for application to datasets of varying sizes and dimensions. Algorithms with lower computational complexity are generally preferable for large datasets or real-time applications, where efficiency is paramount.

By categorizing the algorithms along these dimensions, Table~\ref{table:compmethods} serves as a valuable resource for researchers and practitioners in selecting the most suitable feature selection method for their specific application, balancing methodological approach, learning context, and computational efficiency.

\begin{table}[!t]
\small
\centering
\resizebox{0.48\textwidth}{!}{%
\begin{tabular}{| c |p{2.1cm} |C{0.5cm}| C{0.5cm} |C{2.9cm}|}
\hline
\textbf{ID} &\textbf{Acronym} &   \textbf{\small{Type}} & \textbf{\small{Cl.}} & \textbf{Comp. Complexity} \\\hline
1 & CFS \cite{Guyon:2002} &f&u& $\mathcal{O}(\frac{n^2}{2}T)$  \\\hline
2 & DGUFS\cite{JunGuo_AAAI_2018_DGUFS} & w & u & N/A \\\hline
3 & ECFS ~\cite{RoffoECML16,Roffo2017b} &f&s&  $\mathcal{O}(Tn + n^2)$  \\\hline
4 & Fisher~\cite{Quanquanjournals}   &f&s& $\mathcal{O}(Tn)$   \\\hline
5 & FSASL \cite{du2015unsupervised} & w & u & $\mathcal{O}(n^3 + Tn^2)$ \\\hline
6 & FSV~\cite{Bradley98featureselection} & e& s& \small{$\mathcal{O}(T^2n^2)$} \\\hline
7 & ILFS \cite{roffo2017infinite} &f&s& $\mathcal{O}(n^{2.37}+in+T+C)$  \\\hline
8 & LASSO \cite{liu2007computational} & e & s &\small{$\mathcal{O}(T^2n^2)$} \\\hline
9 & LLCFS \cite{zeng2011feature}  &f&u& N/A  \\\hline
10 & LS \cite{HCN05a} &f&u& N/A  \\\hline
11 & MCFS \cite{Cai:2010}&f&u&  N/A  \\\hline
12 & MI~\cite{Hutter:02feature} &f& s&\small{$\mathcal{O}(T^2n^2)$}\\\hline
13 & Relief-F~\cite{liu2008} &f&s& $\mathcal{O}(iTnC)$  \\\hline
14 & RFE \cite{Guyon:2002} & w& s& \small{$\mathcal{O}(T^2 n log_2n )$} \\\hline
15& UDFS \cite{Yang:2011}  &f&u& N/A \\\hline
16 & UFSOL\cite{guo2017unsupervised} & w & u & $\mathcal{O}(iTCn^3)$ \\\hline
17 & L0\cite{Guyon:2002} &w&s& $ N/A$   \\
18& \textbf{Inf-FS} \cite{Roffo:InfFS:2015} &f&u& $\mathcal{O}(n^{2.37}(1+T))$ \\\hline
19 & mRMR~\cite{Peng05featureselection} &f&s& $\mathcal{O}(n^3 T^2)$\\\hline
\end{tabular}
}
\caption{List of the feature selection approaches provided with the \emph{Feature Selection Library (FSLib)}. The table reports their \emph{Type}, class (\emph{Cl.}), and complexity (\emph{Compl.}). As for the complexity, $T$ is the number of samples, $n$ is the number of initial features, $i$ is the number of iterations in the case of iterative algorithms, and $C$ is the number of classes. The complexity of FSV cannot be specified since it is a wrapper (it depends on the chosen classifier).}
\label{table:compmethods}
\end{table}

\section{Concluding Remarks}\label{sec:conc} 

The work presented in this paper addresses the critical area of feature selection (FS) in machine learning, offering a comprehensive review of methods and introducing the Feature Selection Library (FSLib)\cite{Roffo2020InfiniteFeature,Roffo2019AutomatingAdministration,Vinciarelli2019WeAreLessFree,Roffo2016FeatureSelection,Roffo2017InfiniteLatent,Melzi2018DiscreteTime,Scibelli2018DepressionSpeaks,Roffo2016OnlineFeature,Roffo2019TypeLikeAMan,Roffo2016VOT2016,Roffo2017InfiniteLatent,Roffo2020InfiniteFeature,Vinciarelli2019WeAreLessFree}. This library has been designed to provide researchers and practitioners with a versatile and powerful tool for feature selection, facilitating the enhancement of machine learning models through the efficient identification of relevant features.

The FSLib encompasses a wide array of feature selection algorithms, each catering to different requirements based on the type of data, the desired level of supervision, and computational constraints. By integrating these methods into a single library with uniform input and output formats, FSLib significantly streamlines the process of feature selection, enabling users to compare and apply various methods effortlessly. This uniformity is crucial for conducting large-scale performance evaluations, allowing for the systematic assessment of different algorithms across diverse datasets and application domains.

One of the key contributions of this work is the categorization of feature selection methods within the FSLib according to their operational type—filters, wrappers, embedded methods—and their application context—supervised and unsupervised learning. This categorization not only aids users in selecting the most appropriate method for their specific needs but also provides insights into the underlying mechanisms of each approach. Filter methods, for instance, are invaluable for their computational efficiency and independence from learning models, making them suitable for preliminary feature reduction. Wrapper methods, by leveraging model performance as a criterion for feature selection, offer a more targeted approach but at a higher computational cost. Embedded methods, integrating feature selection within the learning algorithm, present a balanced approach by simultaneously optimizing feature selection and model training.

The library's documentation of computational complexity for each method further enhances its utility, guiding users towards selecting algorithms that align with their computational resources and efficiency requirements. This aspect is particularly relevant in the era of big data, where the volume and dimensionality of datasets continually grow, necessitating efficient and scalable solutions for feature selection.

The public availability of FSLib on MATLAB Central's File Exchange is a testament to the commitment to open science and the democratization of advanced machine learning tools. By providing access to a wide audience, the library not only facilitates the advancement of research in feature selection but also encourages the adoption of best practices in feature selection across various fields of application, from computer vision to bioinformatics.

In summary, the contributions of this work are manifold. Firstly, it provides a structured overview of feature selection methods, elucidating their mechanisms, advantages, and limitations. Secondly, the development and dissemination of FSLib represent a significant practical contribution, offering a tool that simplifies and enhances the process of feature selection. Lastly, by standardizing the interface and output of different feature selection algorithms, FSLib enables comprehensive performance evaluations, fostering a deeper understanding of the efficacy of various methods across different scenarios.

Future work will focus on expanding the library to include emerging feature selection algorithms, especially those leveraging advancements in deep learning and artificial intelligence. Moreover, efforts will be directed towards enhancing the library's scalability and efficiency, ensuring its applicability to increasingly large and complex datasets. By continuing to evolve FSLib, we aim to sustain its relevance and utility in the rapidly advancing field of machine learning, contributing to the ongoing enhancement of model performance and the discovery of insightful patterns in data.

\bibliographystyle{unsrt}

\bibliography{main}  

\end{document}